\documentclass[letterpaper, 10pt, conference]{ieeeconf}
\IEEEoverridecommandlockouts                              

\usepackage[utf8]{inputenc}
\usepackage{graphicx}
\usepackage{amsmath}
\usepackage{amssymb}
\usepackage{url}
\usepackage{xcolor}
\usepackage[justification=centering]{caption}

\title{Forecasting Nonverbal Social Signals during Dyadic Interactions  \\ with Generative Adversarial Neural Networks}

\author{Nguyen Tan Viet Tuyen, Oya Celiktutan
\thanks{}
\thanks{The authors are with the Social AI and Robotics Lab, Centre for Robotics Research, Department of Engineering, King's College London, United Kingdom {\tt\small \{tan\_viet\_tuyen.nguyen, oya.celiktutan\}@kcl.ac.uk}}
}

\begin{document}
\maketitle
\thispagestyle{empty}
\pagestyle{empty}

\begin{abstract}
We are approaching a future where social robots will progressively become widespread in many aspects of our daily lives, including education, healthcare, work, and personal use. All of such practical applications require that humans and robots collaborate in human environments, where social interaction is unavoidable. Along with verbal communication, successful social interaction is closely coupled with the interplay between nonverbal perception and action mechanisms, such as observation of gaze behaviour and following their attention, coordinating the form and function of hand gestures. Humans perform nonverbal communication in an instinctive and adaptive manner, with no effort. For robots to be successful in our social landscape, they should therefore engage in social interactions in a humanlike way, with increasing levels of autonomy. In particular, nonverbal gestures are expected to endow social robots with the capability of emphasizing their speech, or showing their intentions. Motivated by this, our research sheds a light on modeling human behaviors in social interactions, specifically, forecasting human nonverbal social signals during dyadic interactions, with an overarching goal of developing robotic interfaces that can learn to imitate human dyadic interactions. Such an approach will ensure the messages encoded in the robot gestures could be perceived by interacting partners in a facile and transparent manner, which could help improve the interacting partner perception and makes the social interaction outcomes enhanced.
\end{abstract}
\section{Introduction}
This is the fact sheet's for the ICCV 2021 Understanding Social Behavior in Dyadic and Small Group Interactions Challenge~\cite{chalearn:iccv:2021}, ``Behavior forecasting Track''. 
\begin{itemize}
    \item Team name:  SAIR KCL
    \item Username on Codalab: tuyennguyen
    \item Team leader affiliation: Social AI and Robotics Lab (SAIR), Centre for Robotics Research (CoRe), Department of Engineering, King's College London, United Kingdom.
    \item Team leader email: tan\_viet\_tuyen.nguyen@kcl.ac.uk
    \item Name of other team members (and affiliation): \\
    Oya Celiktutan, SAIR, CoRe, Department of Engineering, King's College London, United Kingdom. 
    \item Team website URL: \\
    \url{https://sairlab.github.io}
\end{itemize}

\section{Overview of the Proposed Approach}
\begin{figure*}[t!]
	\centering
	\includegraphics[width= 6.5in]{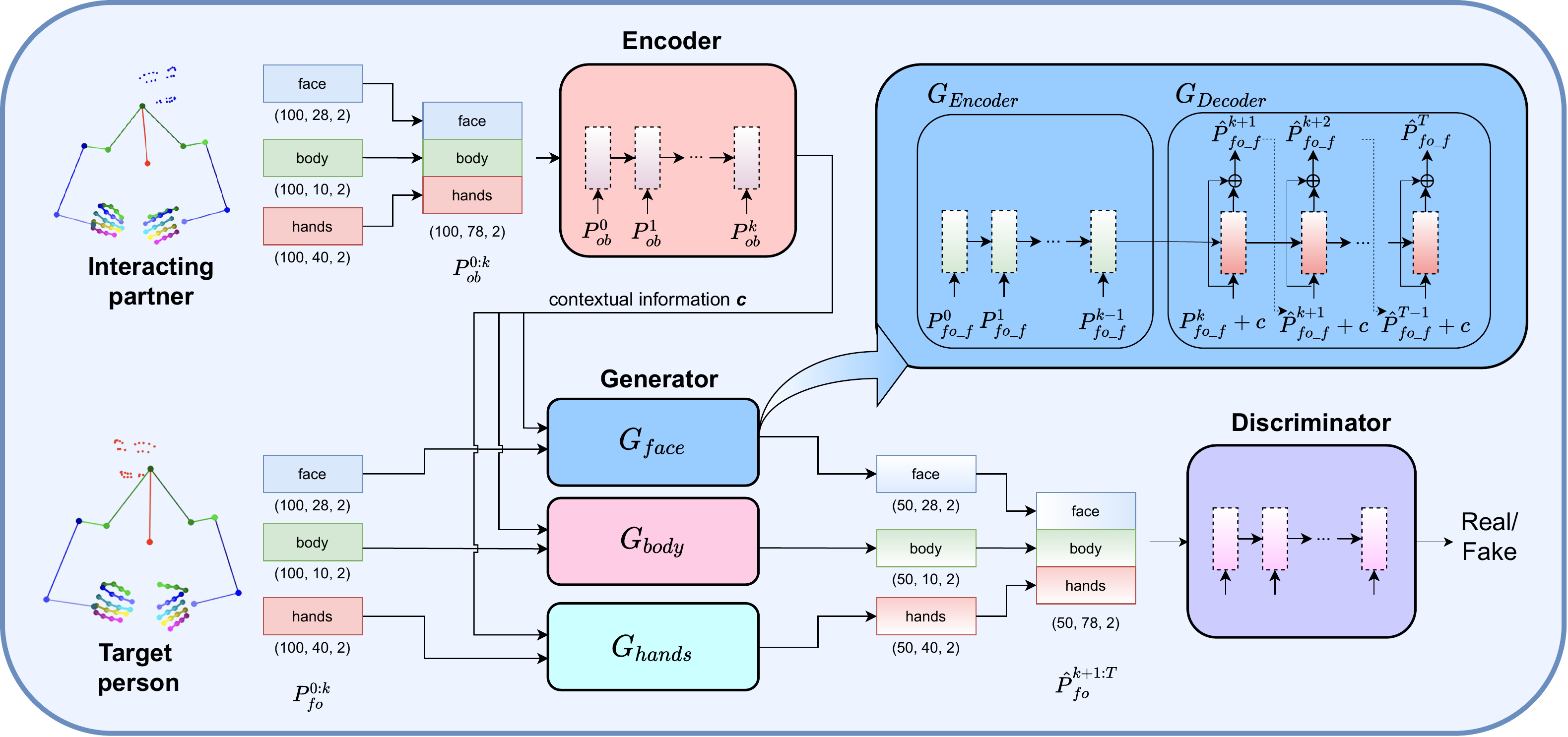}
	\caption{ The designed framework to forecast the non-verbal behaviors $\hat{P}^{k+1:T}_{fo}$ of the target person during socially dyadic interaction. Inputs consist of previous information of the target person $P^{0:k}_{fo}$ and the interacting partner $P^{0:k}_{ob}$.}
	\label{fig:GANmodel}
\end{figure*}

Fig.~\ref{fig:GANmodel} illustrates the proposed training framework to forecast the non-verbal information $\hat{P}^{k+1:T}_{fo}$ $(k+1 \leq t \leq T)$ of the target person during dyadic interaction. The process is started by encoding face, body and hand landmarks $P^{0:k}_{ob}$ $(0 \leq t \leq k)$ of the interacting partner into a contextual vector $c$. It is followed by dividing the non-verbal features of the target person $P^{0:k}_{fo}$ at the same observed time window into three parts, namely face $P^{0:k}_{fo\_f}$, body $P^{0:k}_{fo\_b}$, and hands $P^{0:k}_{fo\_h}$, each of them is injected to a corresponding generator network. By combining with the contextual vector $c$, the three generator networks predict future landmarks of face $\hat{P}^{k+1:T}_{fo\_f}$,  body $\hat{P}^{k+1:T}_{fo\_b}$, and hands $\hat{P}^{k+1:T}_{fo\_h}$. Those features are concatenated into the single vector $\hat{P}^{k+1:T}_{fo}$ representing the predicted motion of the target person. Finally, both of the real gesture $P^{k+1:T}_{fo}$ and the predicted one $\hat{P}^{k+1:T}_{fo}$ are fed to the Discriminator network. The following section will explain the proposed method in more detail.

\section{Detailed method description}
\subsection{Encoder} 
During social human-human interaction, people tend to use a wide range of non-verbal channels to communicate their intentions or emotions~\cite{noroozi2018survey}. Such social signals would influence the other interlocutor's perception and actions, in particular, their non-verbal gestures. In other words, the non-verbal signals of the interacting partner should be treated as the essential stimuli to forecast the target person's motions. Inspiring from that, in the framework presented in Fig.\ref{fig:GANmodel}, a Recurrent Neural Network (RNN) $Encoder$ is introduced to encode the interacting partner's motion $P^{0:k}_{ob}$ consisting of face, body, and hands represented by $2D$ joint coordinates into a vector $c$. The network is designed with a Long-Short Term Memory (LSTM) layer and followed by a fully connected layer to output the fix-length contextual information $c$. Finally, $c$ is fed to the three $Generator$ networks and treated as the conditional information to generate the predicted motion.

\subsection{Generator}
The framework illustrated in Fig.~\ref{fig:GANmodel} is equipped with three $Generator$ networks, namely $G_{face}$, $G_{body}$, and $G_{hand}$. Those are implemented to forecast the motions of face, body, and hands. This strategy allows different motion features could be treated in appropriate manners.  It should be noticed that we address the problem of motion prediction by creating a sequence-to-sequence network including $G_{Encoder}$ and $G_{Decoder}$, they are built upon LSTM layers. $G_{Decoder}$ receives the internal representation encoded by $G_{Encoder}$ and contextual vector $c$ to generate the predicted motion. Noticed that a residual connection is added between the input and the output of each LSTM cell of $G_{Decoder}$, this approach allows the network to better model the velocity of motion~\cite{martinez2017human}. At each time step, instead of employing the ``teacher forcing'' technique, $G_{Decoder}$ receives its own prediction to forecast the next motion frame.

\subsection{Discriminator}
$Discriminator$ is created with a LSTM layer and a fully connected layer to produce an output probability indicating whether the input motion is real or fake. By training $Discriminator$ a capability of distinguishing between $P^{k+1:T}_{fo}$ and $\hat{P}^{k+1:T}_{fo}$, the adversarial loss~\cite{goodfellow2014generative} encourages $Generator$ to produce more realistic motions.  

Overall, the framework demonstrated in Fig.~\ref{fig:GANmodel} is trained with the loss functions defined in Eq.~\ref{eq:loss_G} and Eq.~\ref{eq:loss_D}. We used $\mathcal{L}_G$ to train the $Encoder$ and $Generator$ while $\mathcal{L}_D$ is taken into account for optimizing $Discriminator$. Here, $\mathcal{L}^{MSE}_{f}$, $\mathcal{L}^{MSE}_{b}$, $\mathcal{L}^{MSE}_{h}$ are the mean square errors between the ground truth and the generated motions of face, body, and hands. $\alpha_1$, $\alpha_2$, $\alpha_3$, and $\beta$ are parameters to control the weights of the loss terms.

\begin{align}
\begin{aligned}
\label{eq:loss_G}
\mathcal{L}_G = \alpha_1*\mathcal{L}^{MSE}_{f} & +  \alpha_2*\mathcal{L}^{MSE}_{b} + \alpha_3*\mathcal{L}^{MSE}_{h} \\
      & + \beta*\mathbb{E}\left[ log(1 - D(\hat{P}^{k+1:T}_{fo}) \right]
\end{aligned}
\end{align}

\begin{equation}
\label{eq:loss_D}
\mathcal{L}_D = -\mathbb{E}\left[log(D(P^{k+1:T}_{fo})) \right] -\mathbb{E}\left[ log(1 - D(\hat{P}^{k+1:T}_{fo}) \right]
\end{equation}

\section{Experimental results}
\subsection*{Pre-processing:} For training the framework, we used the training set of the UDIVA v0.5~\cite{palmero2021context}. On each interaction session, annotation data of two interlocutors were pad into 150 frames (6 seconds) ($P^{0:T} \in \mathbb{R}^{150 \times 78 \times 2}$). The first 100 frames (4 seconds) were used as the observed windows, and the last 50 frames (2 seconds) were used as the ground truth motions. As described in Eq.~\ref{eq:normalized}, joint coordinates of the motion frame $P^{i} (0 \leq i \leq 150)$ were normalized taking into account the mean $\mu$ and standard deviation $\sigma$ of the whole motion sequence. Finally, we obtained $30964$ training samples.

\begin{equation}
\label{eq:normalized}
P^{i} = \frac{P^{i} - \mu}{\sigma + 10^{-8}}
\end{equation}

\subsection*{Experimental results:}
The training data was fed to the framework with a batch size of $1024$. The model was trained for $1000$ epochs. We used the Adam optimizer at the learning rate of $5\times10^{-4}$ for $Encoder$, $Generator$, and $Discriminator$. Weights for the loss term were chosen empirically ($\alpha_1=10$, $\alpha_2=10$, $\alpha_3=10$, and $\beta=1$). The adversarial loss mentioned in Eq.~\ref{eq:loss_G} was not used during the first 50 warm-up epochs.

At the testing phase, pairs of motion data collected from the target person $P^{0:k}_{fo}$ and the interacting partner $P^{0:k}_{ob}$ in a period of 4 seconds were sequentially filled in the framework. $Generator$ releases predicted motions of the target person $P^{k+1:T}_{fo}$ in the next 2 seconds. Fig.~\ref{fig:GANmodel} presents a generated action conducted on the testing data. Using the evaluation metric defined by the DYAD challenge organizers, Table~\ref{tab:track2:leaderboard} presents the best result that we obtained on the testing set.

\begin{figure*}[t!]
	\centering
	\includegraphics[width= 6.5in]{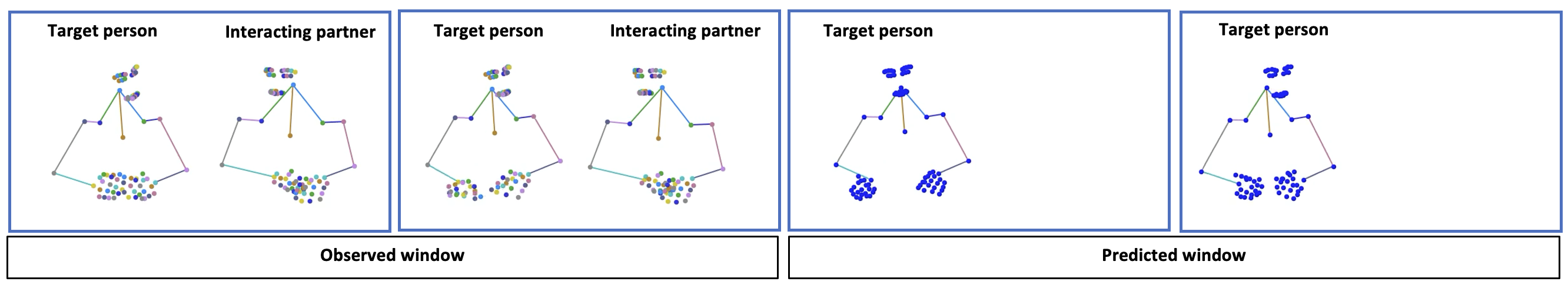}
	\caption{The network receives the non-verbal data of the two interlocutors in a period of 4 seconds and predicts the action of the target person in the next 2 seconds.}
	\label{fig:result_test}
\end{figure*}

\begin{table}[htbp]
\centering
\caption{The best results from Leaderboard (test phase) obtained by the proposed approach.}
\begin{tabular}{|c|c|c|c|}
\hline
\textbf{Avg. Rank} & \textbf{Face} & \textbf{Body} & \textbf{Hands} \\ \hline
3.000 & 0.205 & 0.851 & 0.316\\ \hline
\end{tabular}
\label{tab:track2:leaderboard}
\end{table}

\subsection{Final remarks}
We observed that generated hand motions yielded lower performance than face and body. This problem could be explained by the randomness of hand gestures performed by the target person during social interaction. Such rhythmic movements are well-known as ``beat gesture''~\cite{mcneill2011hand} which are commonly used for stressing keywords or phrases of the communicator's speech. To better approach this motion, the audio features should be integrated into the proposed framework.

\section{Additional method details}

%
%
\begin{itemize}
\item \textbf{Mark with an X the modalities you have exploited.} (~)~Visual, (~)~Acoustic, (~)~Transcripts, (~)~Metadata, (X)~Landmark annotations, (~)~Eye-gaze vectors.\\
%
%

\item \textbf{In case you used metadata, mark with an X the types of metadata you have exploited.} (~)~Age, (~)~Gender, (~)~Country of origin, (~)~Max. level of education, (~)~Pre-session mood, (~)~Pre-session fatigue, (~)~Relationship among interactants, (~)~Task type, (~)~Task order, (~)~Task difficulty, (~)~Language, (~)~Other.\\
If other, or if you have used just a subset of info for a given type of metadata (e.g., just a subset of mood values), please detail: \\
%
%

\item \textbf{Mark with an X the tasks you used for training.} (X)~Talk, (~)~Lego, (~)~Animals, (~)~Ghost.\\
%
%

\item \textbf{Mark with an X the data representation type you used as input.}
(X)~Raw 2D coordinates, (~)~Raw 3D coordinates, (~)~Coordinate offsets (i.e. per-frame displacements), (~)~Velocity, (~)~Acceleration, (~)~Trajectories, (~)~Heatmaps, (~)~Other.\\
Please detail: Annotation data represented by 2D coordinates were pad into 150 frames. Each frame has a shape of (78,2) containing 28 face landmarks, 10 body landmarks, 20 left-hand landmarks, and 20 right-hand landmarks.
%
%

\item \textbf{Mark with an X the data representation type you used as output.}
(X)~Raw 2D coordinates, (~)~Raw 3D coordinates, (~)~Coordinate offsets (i.e. per-frame displacements, (~)~Velocity, (~)~Acceleration, (~)~Trajectories, (~)~Heatmaps, (~)~Other.\\
Please detail: The generated motion $P^{0:k}_{fo}$ has a shape of (50, 78, 2) including 50 motion frames. Each frame has a shape of (78,2) similar as the input.
%

\item \textbf{Did you use information from previous sessions of the target interlocutor as a prior to model his/her behavior for a given (future) session?} (~) Yes, (X) No\\
If yes, please detail:
%
%

\item \textbf{Did you use the minimum observable window of 4 seconds to predict future frames, or another approach?} (X) ``observable window of 4 seconds", (~) ``another approach".\\
If you used a different approach, please detail:\\
%
%

\item \textbf{Did you use the provided validation set as part of your training set?} () Yes, (X) No\\
If yes, please detail:\\
%
%

\item \textbf{Did you use any fusion strategy of modalities?} (~) Yes, (X) No\\
If yes, please detail:\\
%
%

\item \textbf{Did you use the given/predicted personality labels?} (~) Yes, (X) No\\
If yes, please detail:\\
%
%

\item \textbf{Did you treat the face, body and hands as different groups or did you predict all landmarks at once?} (X) ``different groups'', (~) ``all at once''.\\
%
%

\item \textbf{Did you use information from the other interlocutor (e.g., their visual info) to predict the future behavior of the target interlocutor?} (X) Yes, (~) No.\\
If yes, please detail: The motion data $P^{0:k}_{ob}$ of the other interlocutor was treated as a conditional input to forecast the action $P^{k+1:T}_{fo}$ of the target person. 
%
%

\item \textbf{Did you use pre-trained models?} (~) Yes, (X) No\\
If yes, please detail:\\
%
%

\item \textbf{Did you use any face / hand / body landmark detection, alignment or segmentation strategy, instead of or in addition to the landmark annotations provided by the dataset?} (~) Yes, (X) No\\
If yes, please detail:\\
%
%

\item \textbf{Did you use external data?} (~) Yes, (X) No\\
If yes, please detail:\\
%
%

\item \textbf{Did you use any regularization strategies/terms?} (~) Yes, (X) No\\
If yes, please detail:\\
%
%

\item \textbf{Did you use handcrafted features?} (~) Yes, (X) No\\
If yes, please detail:\\
%
%

\item \textbf{Did you exploit depth (i.e., Z component) information?} (~) Yes, (X) No\\
If yes, please detail:\\
%
%

\item \textbf{Did you use any spatio-temporal feature extraction strategy?} (~) Yes, (X) No\\
If yes, please detail:\\
%
%

\item \textbf{Did you use different weights for face, body and hands?} (~) Yes, (X) No\\
If yes, please detail:\\
%
%

\item \textbf{Did you perform any data augmentation?} \\(~) Yes, (X) No\\
If yes, please detail:\\
%
%

\item \textbf{Did you use any bias mitigation technique (e.g., rebalancing training data)?} \\(~) Yes, (X) No\\
If yes, please detail:\\
%
%

\item \textbf{Did you use any input normalization technique (e.g., root-relative coordinates)?} \\(X) Yes, (~) No\\
If yes, please detail: The motion data were normalized/denormalized taking into account its mean and standard deviation values. 
%
%

\end{itemize}

\section{Code repository}
The codes used in this work are available at: \\ \url{https://github.com/TuyenNguyenTanViet/ForecastingNonverbalSignals}
\section{ACKNOWLEDGMENT}

This work has been supported by the ``LISI \-- Learning to Imitate Nonverbal Communication Dynamics for Human-Robot Social Interaction'' Project, funded by the Engineering and Physical Sciences Research Council (Grant Ref.: EP/V010875/1).

\bibliographystyle{IEEEtran}
\bibliography{references}
\end{document}